\documentclass[letterpaper, 10 pt, conference]{ieeeconf}  
\IEEEoverridecommandlockouts                              
\usepackage{cite}
\usepackage{times}
\usepackage{soul}
\usepackage{url}
\usepackage[hidelinks]{hyperref}
\usepackage[utf8]{inputenc}
\usepackage[small]{caption}
\usepackage{graphicx}
\usepackage{amsmath}
\usepackage{amsfonts}

\usepackage{amsthm}
\usepackage{booktabs}
\usepackage{algorithm}
\usepackage{algorithmic}
\usepackage{color}
\usepackage[switch]{lineno}

\title{\LARGE \bf
Recover Triggered States: Protect Model Against Backdoor Attack in Reinforcement Learning
}

\author{Hao Chen$^1$, Chen Gong$^1$, Yizhe Wang$^1$, Xinwen Hou$^1$
\thanks{This work was supported by Chinese Academy of Sciences and University of Chinese Academy of Sciences}
\thanks{Corresponding author: Xinwen Hou, {\tt\small }}%
\thanks{$^1$Institute of Automation, Chinese Academy of Sciences, Beijing, China}%
}

\begin{document}

\maketitle
\thispagestyle{empty}
\pagestyle{empty}

\begin{abstract}
A backdoor attack allows a malicious user to manipulate the environment or corrupt the training data, thus inserting a backdoor into the trained agent. Such attacks compromise the RL system's reliability, leading to potentially catastrophic results in various key fields. In contrast, relatively limited research has investigated effective defenses against backdoor attacks in RL. This paper proposes the Recovery Triggered States (RTS) method, a novel approach that effectively protects the victim agents from backdoor attacks. RTS involves building a surrogate network to approximate the dynamics model. Developers can then recover the environment from the triggered state to a clean state, thereby preventing attackers from activating backdoors hidden in the agent by presenting the trigger. When training the surrogate to predict states, we incorporate agent action information to reduce the discrepancy between the actions taken by the agent on predicted states and the actions taken on real states. RTS is the first approach to defend against backdoor attacks in a single-agent setting. Our results show that using RTS, the cumulative reward only decreased by 1.41\% under the backdoor attack.

\end{abstract}


\section{Introduction}
While deep reinforcement learning (DRL) algorithms have brought significant improvements in efficiency in real-world applications~\cite{Chen2021wide}, a growing body of research has recently focused on the potential threat of backdoor attacks on agents trained through DRL algorithms. Malicious attackers can poison the training data or manipulate the training environment of the agents. Developing effective defenses against backdoor attacks in RL is crucial to ensure the security and trustworthiness of RL-based systems. Backdoor attacks can target RL agents used in critical applications such as autonomous driving~\cite{trojancar,gong2022mind}, resource scheduling~\cite{trojanresouce}, robotic control~\cite{trojanrobot}, and competitive games~\cite{gong2022mind}. Although the threat of backdoor attacks on DRL agents is widespread, the research into developing effective defenses to help agents evade such attacks in RL has been insufficient.

Previous method has proposed to eliminate backdoors in multi-agent scenarios~\cite{trojanSeeker} with a backdoor seeker and memory-loss procedure. But it is unrealistic to use this method in single-agent scenarios, because we can not adjust the observation deliberately as we wish in the single-agent setting. What's more, reverse engineering methods which is designed to locate the trigger in the dataset can not locate the trigger of untarget attacks~\cite{neuralcleanse,trojdrl}. Overall, we need to develop new defense methods specifically for single-agent scenarios. In practice, once a backdoor is injected into an RL agent, it becomes difficult to remove it. Additionally, it can be challenging to determine whether an agent has been implanted with a backdoor if the malicious users do not present triggers. It is inspired by the observation that must manipulate the agent's observation by adding the trigger to activate the hidden backdoor within it. 
Therefore, we have designed a novel defense approach called "Recovery Triggered States" (RTS), which aims to restore the state that has been tampered with the trigger back to its original state. Then, the backdoor can not be activated without the presentation of triggers.

To recover the triggered states, we exploit the use of the dynamics model to build surrogate network to approximate the transit function of environments. Subsequently, we can leverage this network to restore any state to its original state. The agent perform an action $a$ to a certain state $s$, and the environment would transit to the next state $s'$ specified with dynamics of the environment $T$. If the attackers manage to steal the trigger $\delta$ and apply it to the state $s'$, they can present these manipulated states, denoted as $s' + \delta$, to the victim agent in order to activate the backdoor. However, if we have access to the simulation function $T$, we can easily restore the original state by computing $s' = T(s,a)$.  We propose a defense strategy called Recovery Triggered States (RTS), which leverages the consistency of the environment's dynamics to prevent the backdoor attack from being triggered.

Current methods use a single-objective deterministic dynamics model to simulate the environment dynamics, which focuses on the accuracy of predicting state, and is referred to as a single-objective dynamics model. However, when employing a single-objective dynamics model for defending against trigger stage attacks in continuous control scenarios, even small gaps between predicted and real states can cause the policy to execute significantly different actions, resulting in a decline in defense performance. This problem is related to the undesired actions in model inference tasks will cause compounding errors. In a $\mathcal{L}$ length model inference task, the agent get the true state at the beginning state $s_{t0}$, then acting on states predicted by the dynamics model. The dynamics model will get the true state $s_{t0}$, and at rest of time, it will predict states $s^p_{t+1}$, with predicted state $s^p_t$ and policy's action $\pi(s^p_t)$. The environment will get actions of the agent $\pi(s_{t0}), \pi(s^p_{t1}),..., \pi(s^p_{t\mathcal{L}})$. The vulnerability and over-fitting of neural networks causes the policy to execute actions differently when getting predicted states compared to real states.

Compared to traditional dynamics model training that solely focuses on fitting the transition function of environments, we recognize the importance of accuracy in action prediction for the predicted states. Thus, we add an action-related regularization term to the training target of the single-deterministic dynamics model. RTS specifically utilizes a dual-objective dynamics model as the defender, which generates predicted states that conform to the transition dynamics and guide the agent to execute actions similar to those taken in the original states. This approach helps to reduce compounding errors. As shown by our theoretical proof, we can improve the defense performance by utilizing predicted states to guide the agent in executing actions similar to those taken on the original state. Thus, whether in a single or consecutive attack scenario, the dual-objective dynamics model performs better than the single-objective dynamics model.

We conducted extensive experiments to compare the performance of victim agents in different settings, including those where backdoors were not triggered, those where attackers triggered backdoors, and those where the defender protected the input from being corrupted by attackers to trigger backdoors. The results show that the existence of the dual-objective defender can degrade the harm from backdoor attacks, and in average maintain 101.7\% performance of the agent in single-step attack, and maintain 95.49\% performance in consecutive two-steps attack, while the single-objective defender can maintain 86.59\% performance in one step attack and maintain 15.01\% performance in consecutive two steps attack. While agents exposing directly to the backdoor attacks, the performance of the victim agent decreases to 57.57\% and 4.74\%, without the protection of RTS.

We establish the upper bound of the performance of backdoor attacks in this work by their ability to cause deviation of action distributions from those taken on real states, which is caused by the trigger stage attack. The advantages of our method are its ability to resist both target and untarget attacks from being triggered, as well as improving the accuracy of agent's actions on predicted states compared with these taken on real states. In summary, we have developed a defense mechanism against backdoor attacks using a dynamics model, and proposed a dynamics model that is specifically designed for situations involving consecutive attacks. To the best of our knowledge, RTS the first work that utilizes the dynamics model for this purpose. We believe that our work can serve as a source of inspiration for the development of new ideas in the design of dynamics models and abnormal detection techniques. The replication package of RTS are made available online\footnote{\url{https://github.com/Spaxilia/Recover-triggered-states}}.

\section{Related work}
\subsection{Backdoor attacks}
Backdoor attacks can be set in a large number of devices such as a brain-computer interface~\cite{trojanBrain}, and traffic control systems~\cite{trojanTraffic}, code generation~\cite{yang2023stealthy}. Backdoor attacks can also be set in different types of neural networks such as a transformer model~\cite{trojanTransformer}. The backdoor can be activated through imperceptible media such as ultra-sonic sound wave~\cite{trojanUltrasonic}. A backdoor attacker can set different triggers so that he could reinforce his attack ability~\cite{Trojanmulti}. Researchers introduced a backdoor attack method to attack RL agents in single-agent scenarios in electronics games~\cite{trojdrl}. In real world, researchers used backdoor attack to demolish reinforcement-learning-augment autonomous driving. In competition scenarios, researchers have proposed to attack RL agent competing with other agents~\cite{PRE-trojancompetition,chen2022curiosity} and trigger the backdoor through taking a consecutive legal actions allowed in the environment. Researchers also introduced a way to enhance the effectiveness of the backdoor attack~\cite{gong2022mind}.

These backdoor attack methods have threaten the use of the intelligent agent in different aspects of our society. We should not only highlight the security issue on data but also design algorithms for defending against backdoor attacks.

\begin{figure}
    \centering
    \includegraphics[width=1\linewidth]{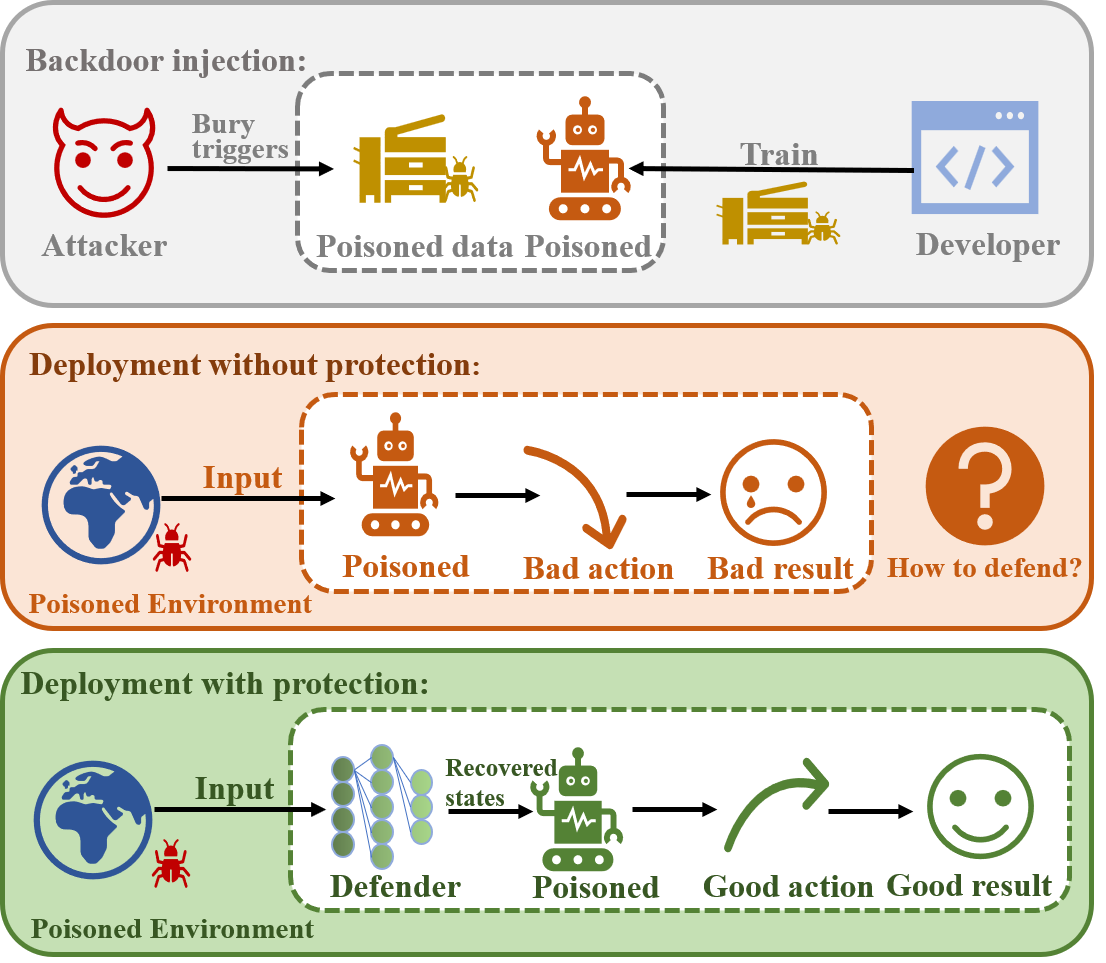}
    \caption{Our threat and defense model involves backdoor attackers leaving poisoned datasets online or burying triggers in unprotected datasets. RL developers then unknowingly train agents using these datasets, resulting in the agents being injected with backdoors. The poisoned agents may still perform well in the developers' view, but when the attacker presents triggered states to the victim agent, its performance deteriorates. To defend against this threat, our model uses a defender to recognize abnormal inputs and provide a predicted state to guide the agent, thereby preventing it from executing an abnormal action.}
    \label{fig:TDPipeline}
\end{figure}

\subsection{Defending backdoor attacks for RL}
Many of methods defending against backdoor attacks need true samples, such as \cite{defentrusam1,defentrusam2}. But true samples can not be provided in most cases of RL, which means some methods used in defending against backdoor attacks in tasks such as picture classification cannot be used in RL. Researchers also introduced a backdoor seeker and memory-loss technique to eliminate backdoors in multi-agent competition scenarios \cite{trojanSeeker}, and it is not applicable as mentioned above. And people have tried to use reverse engineering \cite{neuralcleanse} to locate the polluted data before training the agent, but it didn't work well in locating the untarget attack because random neurons are activated as shown in \cite{trojdrl}.

Recently, researchers also introduced defense methods that don't need true sample in classification tasks, such as using sparsity \cite{sparsity}, which may help defend against backdoor attacks in RL.

\section{Preliminaries}

The purpose of the deep reinforcement learning (DRL) algorithm was to train an agent that can solve sequential decision-making tasks modeled as a Markov decision process (MDP). An MDP is formally described as a 4-tuple $\langle \mathcal{S}, \mathcal{A}, \mathcal{R}, \mathcal{T} \rangle$, where $\mathcal{S}$ is the state space, $\mathcal{A}$ is the action space, $\mathcal{R}$ is the reward function, and $\mathcal{T}$ is the transition function.

In DRL, an agent interacts with an environment over a series of time steps. At each time step, the agent observes a state $s_t$ and selects an action $a_t$ based on its policy $\pi(\cdot|s_t)$. The agent then receives an immediate reward $r_t$ from the environment based on its action, and the environment transitions to a new state $s_{t+1}$ as determined by the transition function $\mathcal{T}(\cdot|s_t,a_t)$. This process continues until the agent reaches a termination state $s_T$, generating a trajectory $\mathcal{N}: (s_0,a_0,r_0, s_1,a_1,r_1,\cdots, s_{T}, a_{T}, r_{T}).$ The cumulative discounted return $R(\mathcal{N}) = \sum_{i=0}^T g^i r_i$ is the sum of discounted rewards along the trajectory, where the discount factor $g \in (0,1)$. The agent's goal is to learn an optimal policy $\pi^*$ that maximizes the expected cumulative discounted return over all possible trajectories,
\begin{equation}
\pi^\ast = \arg\max_\pi \mathbb{E} \left[R(\mathcal{N}_\pi) \right],
\label{eq:rl_target}
\end{equation}
where $\mathcal{N}_\pi$ denotes the distribution of trajectories generated by policy $\pi$.

When an agent is implanted with a backdoor, it behaves normally until it encounters a specific \textit{trigger} $\delta$ in its observation, at which point it fails or behaves abnormally. The trigger can be a specific pattern or input the agent has been trained to recognize as a signal to perform a specific action. We formalize the target of a backdoor attack as follows,
\begin{equation}
    \max \mathbb{E} \left[R(\mathcal{N}_{\pi_b}) \right] + \min  \mathbb{E} \left[R \left((\mathcal{N} + \delta \right)_{\pi_b}) \right] 
\end{equation}

where $\pi_b$ refers to the policy that has been implanted with a backdoor, while $\tau + \delta$ denotes a trajectory that includes a state in which the backdoor trigger is present.


\section{The Recovery triggered states Method}
\subsection{The Threat and Protection Model}

In our threat model, a backdoor attack can be separated into two stages, the injection stage and the trigger stage. In the injection stage, the attacker would implant backdoors into the dataset, and the RL trainer will use the polluted dataset to train the model, which will cause the agent being poisoned, as shown in Fig. \ref{fig:TDPipeline}. The proportion of polluted data in the dataset is very low \cite{trojanrobot, trojdrl} thus decreasing the possibility of being detected by human inspection. Backdoor attacks on RL are notorious because it can hardly be removed once injected. And human inspections are impossible to achieve because the amount of data needed for using an offline RL algorithm to train a RL agent is too big, for example, in general we need 10M - 20M pieces of data to use an offline RL algorithm to train a RL agent in the environment of our experiment. The injection stage attack warms us to watch out the data we used for training the agents. In the trigger stage attack, attacker can set one step attack and consecutive attacks, and the later can further deviate the actions of victim agents from their normal values, which makes a severer damage.

In our defense model, we propose to defend against backdoor attacks in the trigger stage. Before deployment, we let the victim agents interact with clean environment, and collect trajectories to train the dynamics model, then apply the dynamics model when the victim agent is deployed for working. Our defense method is usable because the amount of data needed for training the dynamics model is relatively small, for example, in our experiment we need 50K pieces of data to train a dynamics model for single-step state prediction with a loss below 4 in L2 norm. With little effort we could prevent the agent from being harassed by the backdoor attacks, and if we use RL evaluation methods to include much of the circumstances before deployment, we could further improve the defense performance of our RTS method.

Overall, our method defends the attack in a unison and easy way that the defender can detect and prevent the trigger stage attack using the same dynamics model, and protect the victim agent from the target attack and the untarget attack. The tenet for our defence strategy can be concluded as below.
   \begin{eqnarray}\label{defendequation}
   \mathbb{E} \left[R(\mathcal{N}_{\pi_b}) \right] = \mathbb{E} \left[R \left(T^d(\mathcal{N} + \delta )\right)_{\pi_b}) \right]
   \end{eqnarray}

In the equation (\ref{defendequation}), the $T^d$ is our defender. If an attacker wants to trigger the backdoor, he must distort the input and this will cause the discontinuity of the environment dynamics especially in a continuous control scenario thus our defender would easily capture the difference. 

In image-input scenarios, it would be hard to distinguish if the coming states is triggered merely comparing the predicted state and the coming state. In this case, we could use the difference between policy's actions on the predicted state and that on the coming state to determine if the states is triggered, and our proposed action-related regularization term is more welcomed because $\pi(s_p)$ would be similar to $\pi(s)$, thus improving the accuracy of detection a trigger stage attack. 

Overall, the attack performance of backdoor attack is bounded by both whether the attacker can successfully trigger the backdoor and its ability to change the action with the use of triggered input $s^*$, and our RTS method can provide an easy and efficient way to detect the abnormal states and prevent attacks.
\subsection{The Design and Use of the Defender}
\begin{algorithm}[tb]
    \caption{The RTS Method for Defending Against the Backdoor Attack in Continuous Control}
    \label{defense Process}
    \textbf{Input}: Victim Agent $\pi_V$, Dynamics Model $T$, Environment env, Episode Length N, Final State $s_F$, Reward r, Done done, Threshold H.
    \begin{algorithmic}[1] 
        \STATE $s^{(0)} = env.start$
        \STATE $n=0$
        \WHILE{n\textless N or Not done}
        \STATE $a_V^{(n)}$ = $\pi_V(s^{(n)})$
        \STATE $s^{(n+1)}, r, done = env.step(a_V^{(n)})$
        \STATE In this time, the attacker may skew the $s^{n+1}$ into a trigger $s^*$
        \STATE $s^p=T(s^{(n)},a_V^{(n)})$.
        \IF{$\left\|s^p-s^{(n+1)}\right\|_2>H$}
        \STATE $s^{(n)}=s^p$
        \ELSE
        \STATE $s^{(n)}=s^{(n+1)}$
        \ENDIF
        \STATE $n+=1$
        \ENDWHILE
        \STATE \textbf{return}
    \end{algorithmic}
\end{algorithm}

\textbf{Model choice:}
In designing the structure of the defender, considering the data expenditure, we use the single-deterministic model to simulate the environmental dynamics. In the experiment, the error between the predicted state and the real state given by the model is within an L2 norm of 4.

\textbf{Data collection:}
In a clean environment, we let the poisoned agent perform tasks and collect the trajectories data. This premise can be easily fulfilled, because we don't need backdoors to be triggered to collect the data, which can save the time used in data collection. 

\textbf{Train the Defender:}
After data is collected, we begin to train a defender. The target of training the defender can be denoted as  (\ref{dualdynamicstarget}), in which $T^d$ is the defender, $s_{t-1},a_{t-1}$ are the input for the defender, and the $s_t$, $a_t$ are the target.

\begin{eqnarray}
\label{dualdynamicstarget}
\begin{aligned}
T^d=\underset{T(\ldots;\theta)}{\operatorname{argmin}}&\left\|s_t-T\left(s_{t-1}, a_{t-1};\theta\right)\right\|_2+\\&\left\|a_t-\pi(T\left(s_{t-1}, a_{t-1};\theta\right))\right\|_2
\end{aligned}
\end{eqnarray}

\textbf{Defend the Attack:}
At time step $t$, the defender gets previous input $s_{t-1}$ and agent’s action $a_{t-1}$ to give a prediction $s^p$. The state $s_t$ should be close to the state $s^p$. The attacker may distort a state into a triggered state $st^*$. The defender will consider the difference of the $s_t$ and $s^p$. We could measure the difference of other targets, such as the difference between the $\pi(s_t),\pi(s^p)$. If the difference is out of expectation $h$, the defender will proceed to recover the state. Here we use the l2 norm in the experiment as shown in the \ref{detect}.
\begin{eqnarray}
\label{detect}&\left\|T^d(s_t^p \mid s_{t-1}, a_{t-1})-s_{t}\right\|_2>h
\end{eqnarray}

If the defender captures the abnormal distance between the $s_{t}$ and $s_t^p$, the defender would recover a prediction $s_t^p$ which the victim agent acts on, finally stopping the backdoor to be triggered. The neural networks kind dynamics models have a better ability to capture the dynamics in an episodic manner \cite{GlobalOptimalControl, lambert2021learning}, which encourages us to use the model to simulate the dynamics of a fixed policy $\pi$ interacting with the environment.
\subsection{Defense performance bound and its similarity with model inference tasks}
The process of the trigger stage attack of backdoor attacks is similar to that of adversarial attack. In both kinds of the attack, attackers will skew the input of an victim agent thus inducing the victim agent to perform another action. Both of attacks will not influence the dynamics of the environment. Thus, we can use the performance bound of the adversarial noise attack to define the attack performance of the backdoor attacks.

\begin{figure}
    \centering
    \includegraphics[width=\linewidth]{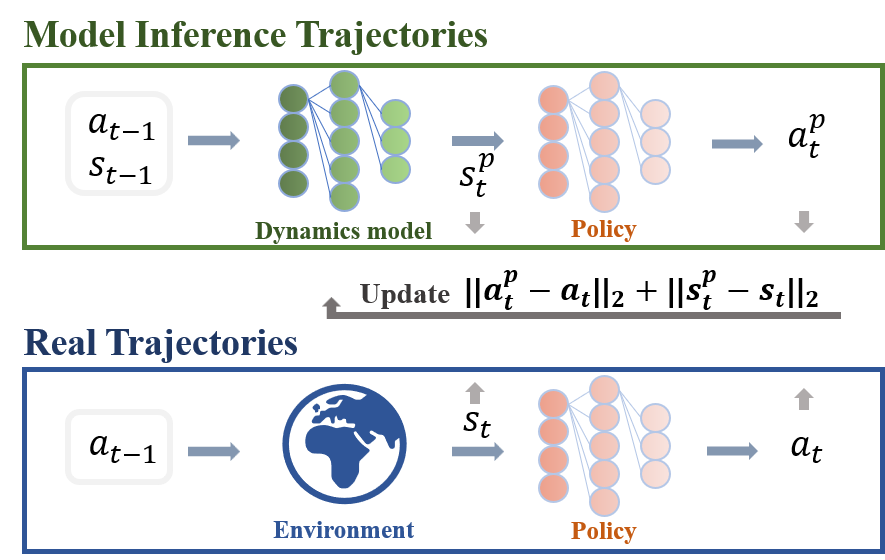}
    \caption{Dual-objective training target of our defender. By adding an action related-term, we improve the dynamics model's ability to predict a state on which the agent will take actions as those taken in real states.}
    \label{fig:dual-gradient}
\end{figure}

Previously, people have defined a performance bound of an adversarial attack as shown in (\ref{advbound}) which shows that an adversarial attack can be determined by its ability to change the action of the policy on the noised input. This equation also indicates an upper limit on the maximum difference between value functions with distinct action distributions \cite{vulnerabilitypolicy}.
\begin{equation}\label{advbound}
\begin{aligned} 
&\max _{s \in S}\left\{V_{\pi}(s)-V_{\pi}(s^*)\right\} \leq \\ &\alpha \max _{s \in S}\max _{s^* \in \beta(s)} D_{TV}(\pi(\cdot| s), \pi (\cdot| s^*))
\end{aligned}
\end{equation}

In Equation (\ref{advbound}), $V_\pi(\cdot)$ is the value function of the policy $\pi$, and $\beta(\cdot)$ outputs $s^*$, state added adversarial noise, and $D_{TV}(\pi(\cdot| s), \pi (\cdot| s^*))$ is the total variation distance between $\pi(\cdot| s)$ and $\pi (\cdot| s^*)$ and $\alpha$ is a constant that does not depend on $\pi$.

As explained above, because these two attack methods share certain similarities, we use (\ref{backbound}) to establish a limit on the effectiveness of the backdoor attack in terms of its capacity to sway the victim's policy through the use of the triggered input $s^*$.

\begin{equation}\label{backbound}
\begin{aligned}
   &\max _{s \in S}\left\{V_{\pi^*}(s)-V_{\pi^*}(s^*)\right\} \leq \\ &\alpha \max _{s \in S}\max _{s^* \in T(s)} D_{TV}(\pi^*(\cdot| s), \pi^* (\cdot| s^*)) 
\end{aligned}
\end{equation}

In Equation (\ref{backbound}), $\pi^*$ means the poisoned agent, and $T(\cdot)$ outputs $s^*$, the state triggered by the attacker, and $D_{TV}(\pi^*(\cdot| s), \pi^* (\cdot| s^*))$ is the total variation distance between $\pi^*(\cdot| s)$ and $\pi^* (\cdot| s^*)$.

Thus, as shown in (\ref{defenderbound}), we can evaluate the defender's performance in preventing the trigger stage attack by examining the differences between $\pi(\cdot| s_{t})$ and $\pi (\cdot| T^d(s_{t-1},a_{t-1}))$.

\begin{equation}\label{defenderbound}
\begin{aligned}
    &\max _{s \in S}\left\{V_{\pi^*}(s)-V_{\pi^*}(T^d(s^*))\right\} \leq \\ &\alpha \max _{s \in S}\max _{s^* \in T(s)} D_{TV}(\pi^*(\cdot| s), \pi^* (\cdot| T^d(s_{t-1},a_{t-1})))
\end{aligned}
\end{equation}

This equation shows the importance of a defender's ability to generate a predicted state and guide a victim agent's actions to closely approximate those in the true state, in order to effectively defend against attacks. Once the defender detects a trigger-stage attack, it supports the victim agent in acting on nearly normal input, rather than maliciously controlled input, to prevent the attack.

We aim to use a single-deterministic dynamics model as a defender to get the advantage of the generalization ability of neural networks as well as avoiding the huge data consumption associated with recurrent neural networks. The training target of the traditional single-deterministic dynamics model is as shown in (\ref{dynamicstarget}),
\begin{eqnarray}
\label{dynamicstarget}
&T^d=\underset{T(\ldots;\theta)}{\operatorname{argmin}}\left\|S_t-T\left(s_{t-1}, a_{t-1};\theta\right)\right\|_2
\end{eqnarray}
which is very common. However, in continuous control scenarios, even gaps between predicted states and real states are small, the policy will execute very different actions.

The problem can be illustrated in (\ref{policydiff}). 
\begin{equation}\label{policydiff}
\|\pi(s)-\pi(s+\Delta)\|_2 \geq L \text {, where } \Delta \leq d
\end{equation}
This issue will hinder the defense performance, which is severer in consecutive attack scenarios. 

This problem is related to the undesired actions in model inference tasks causing accuracy degradation. When policy gets a predicted state different from the real state, it executes an action different from that taken in a real state. And this action further causes the dynamics model generating states different from the real states even if the dynamics model can accurately predict states under a certain bound, and finally accumulates to big compounding errors. Thus, one way to improve the model's inference task performance and reduce the effect brought by the compounding error is to generate a state that guides the agent executing an action similar to that taken on the real state, then the dynamics model could predict accurately. The existence of the adversarial noise \cite{vulnerabilityvalue, vulnerabilitypolicy} reminds us that with little turbulence on the input, the output of the neural networks could be very different. So it is reasonable to use this regularization term in single-deterministic dynamics model as well as other dynamics model to reduce accumulated errors. 

\begin{equation}\label{predictiveissue}
    s_{t+1}^p=f_\theta\left(s_t^p, \pi(s_t^p)\right)
\end{equation}

\begin{figure}
    \centering
    \includegraphics[width=\linewidth]{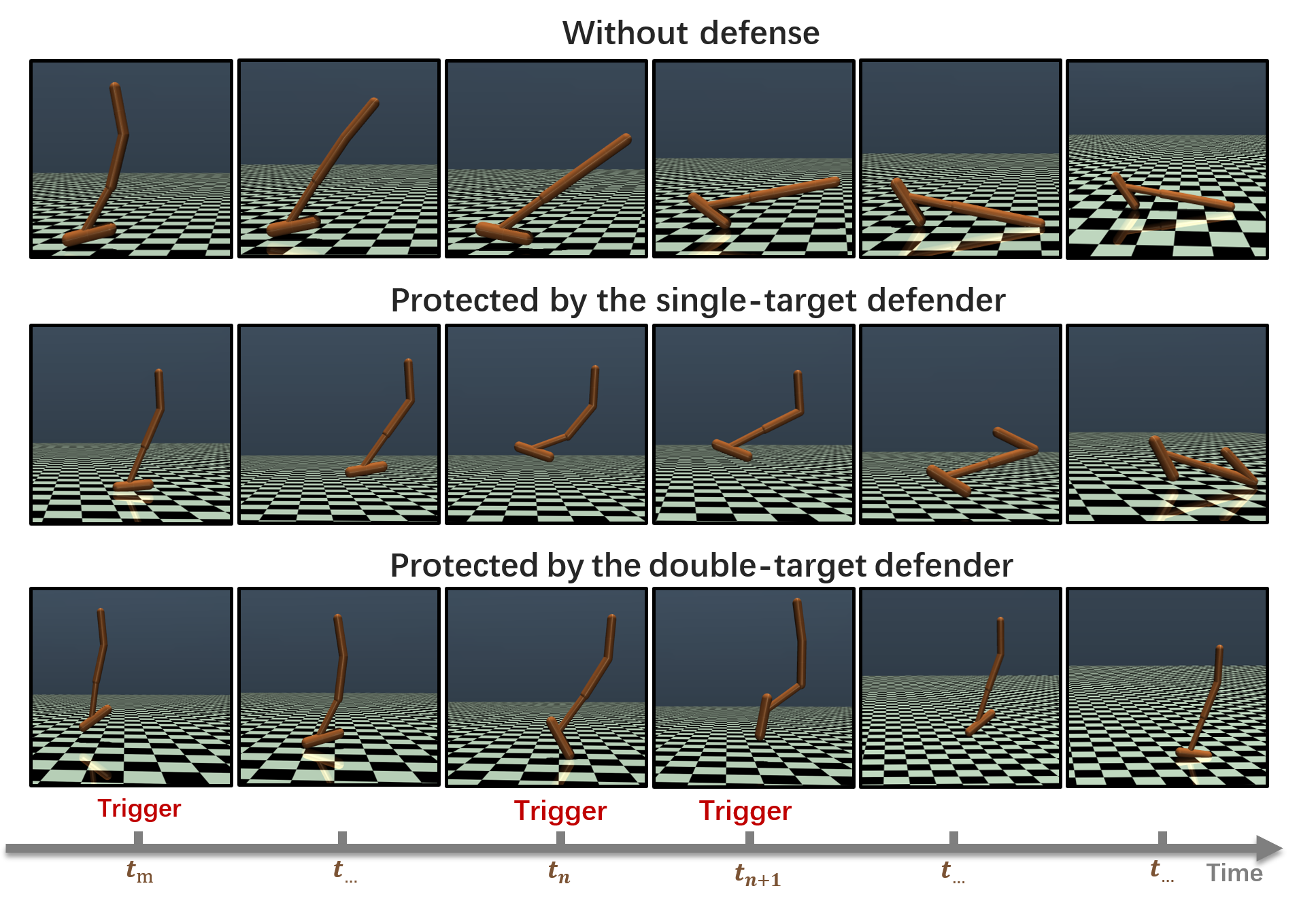}
    \caption{A comparison on different conditions under single attack at $t_m$ and consecutive attacks at $t_n$ and $t_{n+1}$. When we deploy poisoned agents without protection, attackers can easily make the agent fail. When we deploy agents protected by the single-objective defender, the defender can provide protection against single-step attack, but fails to protect the agent in a consecutive attacks due to compounding errors. When we deploy agents protected by the dual-objective defender, it can protect the poisoned agent from single-step and consecutive steps of trigger stage attack.}
    \label{threecomparisons}
\end{figure}

A dual-objective dynamics model could perform better in both single and consecutive attack scenarios as shown in Fig. \ref{threecomparisons}. The defense performance of the single-objective dynamics model degrades because agent doesn't act as expected and the error further causing the dynamics model to predict an undesired states, finally causing the failure of the agent. And the agent plays to the end with the dual-objective dynamics model as a defender.

By guiding the agents to execute actions as if on a real state, the defender can achieve a smaller distance of $D_{TV}(\pi^*(\cdot| s), \pi^*(\cdot| T^d(s^*)))$, which improves their performance against trigger stage attacks, as shown in (\ref{defenderbound}).

\section{Experiments}

\begin{figure}
    \centering
    \includegraphics[width=\linewidth]{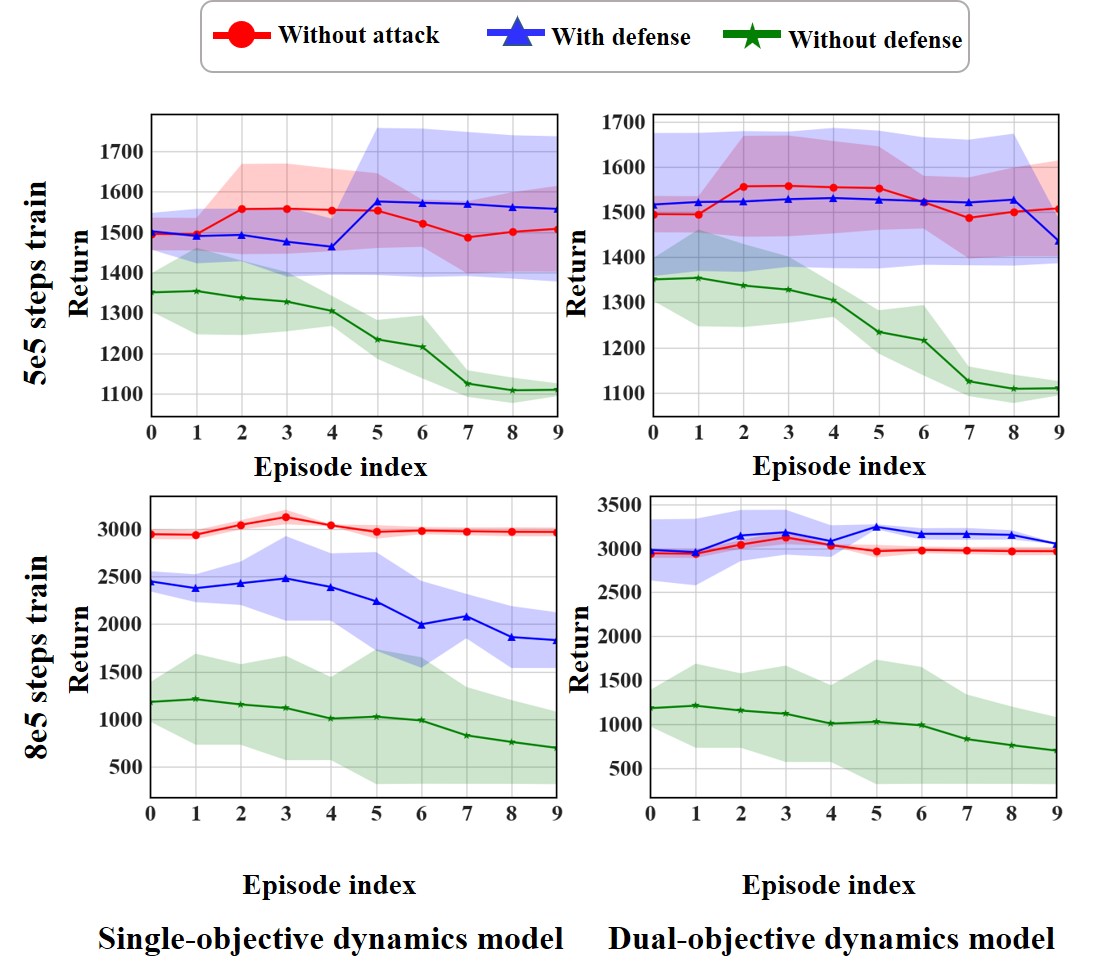}
    \caption{Comparisons in the single step attack. We can see that in the second scenario, the dual-objective dynamics model outperforms the single-objective dynamics model.}
    \label{attack1}
\end{figure}
\begin{figure}
    \centering
    \includegraphics[width=\linewidth]{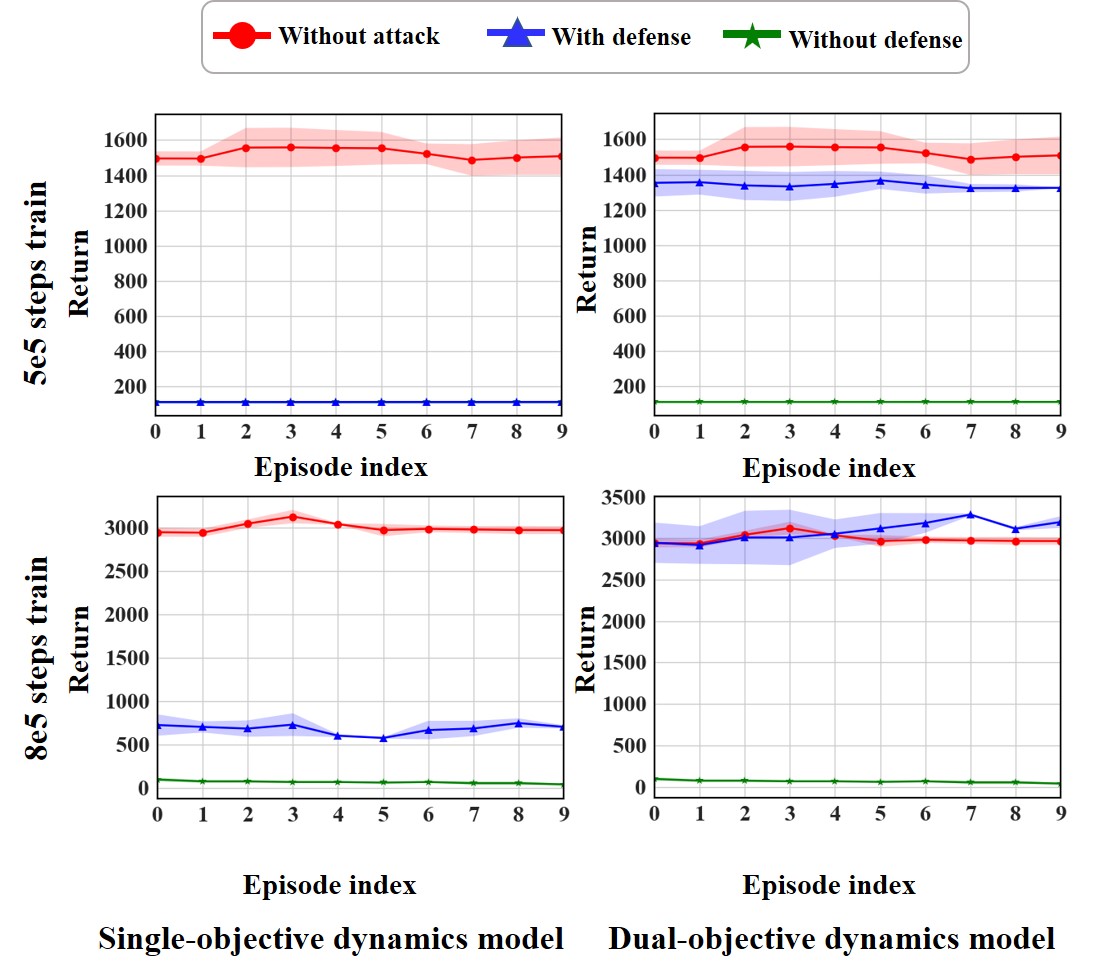}
    \caption{Comparisons in the consecutive two-step attack show that our dual-objective dynamics model outperforms the single-objective model.}
    \label{attack2}
\end{figure}

The Fig. \ref{attack1}, and the Fig. \ref{attack2} are the result of our experiments, where the x-axis indicates the index of the episode and the y-axis indicates the total return of the policy. We test the victim agent trained on the mujoco Hopper environment, and we pick up the victim agents' policy in two time spots of the training on the Hopper, accordingly at 500K steps 800K steps.

\subsection{The defender model's structure}
We use data collected when the victim agent interacts with environments to train the defender. A Defender’s network is: (State Dimension + Action Dimension, 256) (256,256) (256, State Dimension). When collecting the data, we modify the policy through adding a Gaussian noise on its action to maintain exploration. We find with a 0.01 probability to use the noise is sufficient for sampling the data. We use 50K pieces of data to train a defender in the Hopper environment. 
\subsection{The single-step attack}
We use the deep deterministic policy gradient (DDPG) \cite{ddpg} to train the victim model, and inject the target backdoor attack data after the policy has run for 25K steps to reappear the injection stage attack. The proportion of polluted data is 4\% in a dataset of 25K steps, 0.2\% in a dataset of 500K steps, and 0.0125\% in a dataset of 800K steps. 

In the single-step attack scenario, we trigger the backdoor one time every twenty steps by distorting the input when a well trained victim agent is working as shown in \ref{attack1}. On average, the victim agent scores 1523.09 and 2994.43 points without attack in two scenarios. In contrast, the scores decrease to 1246.96 and 998.28 with trigger stage attack. The scores increase again to 1526.24 and 2213.60 when protected by the single-objective defender and to 1515.84 and 3112.66 when protected by the dual-objective defender.

As the agent gradually learns the optimal strategy and executes more steps in a trajectory, the dual-objective defender provides better defense.
\subsection{The consecutive attack}
Then we perform the attack two consecutive steps in every twenty steps, and results as shown in Fig. \ref{attack2}. With trigger stage attack, the scores decrease to 112.08 and 63.84. And the scores increase to 112.89 and 679.38 when protected by the single-objective defender and to 1340.88 and 3084.50 when protected by the dual-objective defender. 

Our dual-objective defender provides a better defense performance compared with that of the single-objective defender.
Due to the existence of perturbations, the scores of the protected agents in attack scenarios are slightly higher than those of the agents who have not been attacked, and this deviation is acceptable.
\begin{figure}
    \centering
    \includegraphics[width=\linewidth]{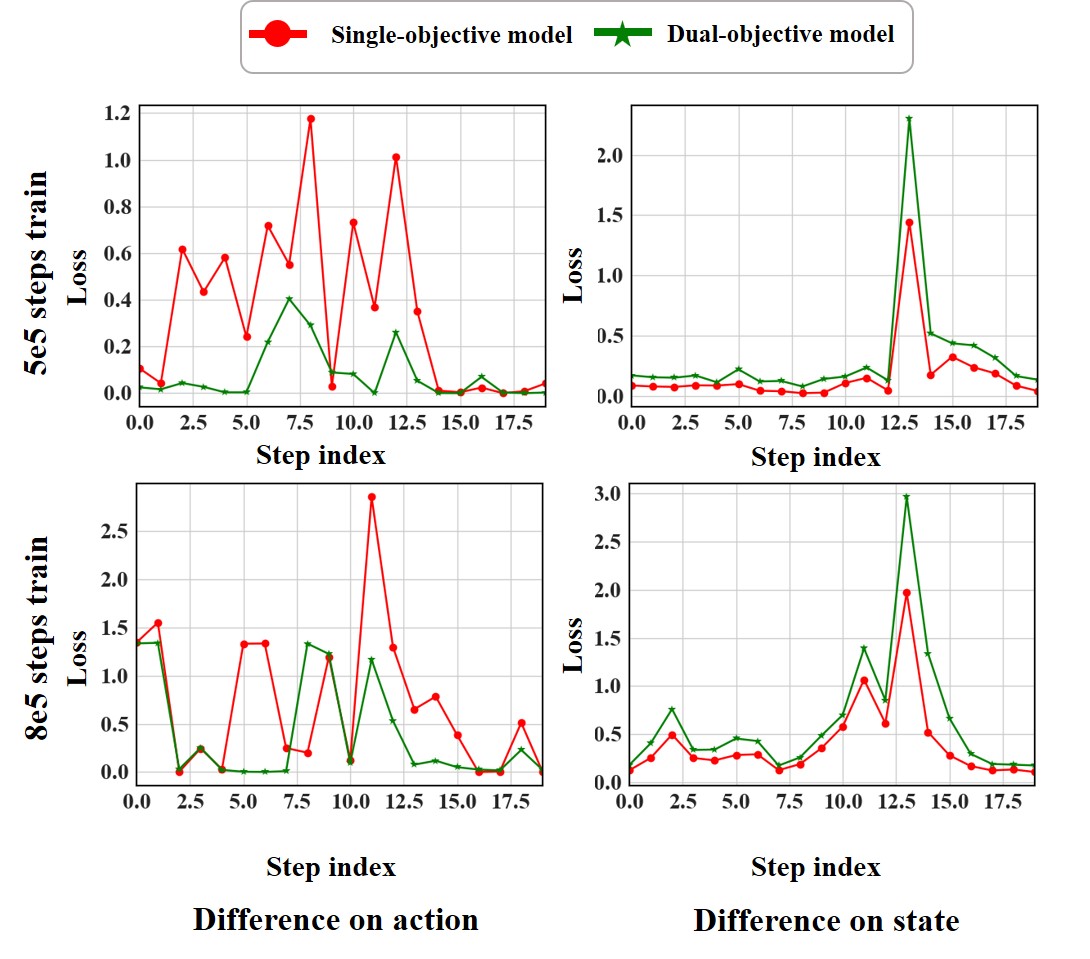}
    \caption{Comparisons of action loss and state loss using different models. In general, on the states predicted by the the dual-dynamics model, agent's actions will be more similar to that taken on a states. However, the states predicted by a single-objective dynamics model are more similar to real states. }
    \label{fig:SAD}
\end{figure}
\subsection{The states loss and action loss}
To verify that the dual-objective dynamics model can generate state guiding agents to execute actions similar to that on real states, we demonstrate the difference between predicted states and real states in a length $\mathcal{L}=1$ model inference task. The predicted state are separately predicted by the single-objective model and the dual-objective model.

An agent trained after 500K reported that, on average, the L2 distance between the real state and the states predicted by the single-objective defender and the dual-objective defender were 0.1715 and 0.3125, respectively. Additionally, the average L2 distance between actions on predicted states and real states were 0.3525 and 0.0796.

An agent trained after 800K reported that, on average, the L2 distance between the real state and the states predicted by the single-objective defender and the dual-objective defender were 0.4099 and 0.6304, respectively. Additionally, the average L2 distance between actions on predicted states and real states were 0.7043 and 0.3941.

The results are as shown in Fig. \ref{fig:SAD}. These result confirm that our dual-objective defender can predict states guiding the agent to act as if on real states.

\section{Discussion AND FUTURE WORKS}
The experiments results also confirm our theorem that defense method will be better if agent's actions on the predicted states are closer to the actions executed on real states by the agent, thus the dual-objective dynamics model performs better than the single-objective dynamics model as a defender against trigger stage attacks. 

We also want to introduce our regularization to other dynamics models and finally use them in the model-based RL. Like the long-gap-term prediction model trained with episodic data in \cite{lambert2021learning}, we train a dual-objective model with data generated by a fixed policy interacting with the environment, so we are facing an associated question that can the episodic-related dynamics model encourage a better performance in model-based RL. Whether our regularization can be used in improving the model-based reinforcement learning is waiting to be revealed. Moreover, it is possible to apply the action-related regularization term to other types of dynamics models, but if it is effective in such cases remains to be researched.
\section{CONCLUSIONS}
Our method can degrade backdoor attacks by applying the dynamics model for detecting and defending against trigger stage backdoor attacks. Current works about designing the trigger ignore the criteria that whether it breaks the consistency held in the environment and our work points out the leak, and it is an issue worthy researching how to plant the backdoor without being detected by the dynamics model. We discuss the similarities of adversarial attacks between backdoor attacks and give a performance bound to the backdoor attacks. We present a dual-objective dynamics model can leverage the the compounding errors brought by inaccurate prediction on states. To the best of our knowledge, this is the first work to use a dynamics model for defending against backdoor attacks, as well as the first to introduce an action-related regularization term in the training of a single-deterministic dynamics model, which we hope would provoke new ideas in the usage of the dynamics model in fields like anomaly detection and model inference task.

\bibliographystyle{IEEEtran}
\bibliography{sample}
\end{document}